\documentclass[conference]{IEEEtran}
\usepackage{times}

\usepackage[numbers]{natbib}
\usepackage{multicol}
\usepackage[bookmarks=true]{hyperref}
\usepackage{graphicx}
\usepackage{tcolorbox}
\usepackage{amsmath}
\usepackage{amssymb}
\usepackage{xcolor}
\usepackage{multirow}
\usepackage[table]{xcolor}
\usepackage{longtable}
\usepackage{geometry}
\usepackage{adjustbox}
\usepackage{subcaption}
\usepackage{booktabs}
\usepackage{CJK}
\usepackage{algorithm}
\usepackage{algorithmicx}
\usepackage{algpseudocode}
\floatname{algorithm}{algorithm}

\definecolor{lightgray}{gray}{0.9}
\definecolor{mydarkblue}{rgb}{0,0.08,0.45}
\definecolor{mydarkgreen}{RGB}{0, 139, 69}
\definecolor{mygreen2}{RGB}{0 205 0}
\definecolor{mybrown}{RGB}{139 69 19}
\definecolor{mypink}{RGB}{184 131 211}
\definecolor{myblue2}{RGB}{187 151 39}
\definecolor{boxblue}{RGB}{79,173,234}
\definecolor{boxgreen}{RGB}{159,206,99}
\definecolor{tablepeach}{RGB}{251, 240, 217}
\definecolor{tablepurple}{RGB}{248,235,252}
\definecolor{tableblue}{RGB}{235,241,255}
\definecolor{lowerbody}{RGB}{76,123,49}
\definecolor{upperbody}{RGB}{47,110,186}

\definecolor{lightblue}{HTML}{7FB2D3}
\definecolor{lightgreen}{HTML}{90B56D}  
\definecolor{darkblue}{HTML}{367DB0}  
\definecolor{darkgreen}{HTML}{3D9F3C} 

\hypersetup{
    colorlinks=true,
    breaklinks=true,
    citecolor={green!70!black},
}

\begin{document}

\title{X-DiffVLA: X-Embodied Diffusion Action Heads for Vision-Language-Action Models}

% You will get a Paper-ID when submitting a pdf file to the conference system
% \author{Author Names Omitted for Anonymous Review. Paper-ID [561]}

\author{%
  \textbf{Boyu Li$^{1,2,3}$, Chaoyi Xu$^{4}$, Haoqi Yuan$^{4,5}$, Xinrun Xu$^{2}$,}\\
  \textbf{Börje F. Karlsson$^{3}$, Haoran Li$^{1,2,3}$, Zongqing Lu$^{4,5}$, Dongbin Zhao$^{1,2,3\authorrefmark{1}}$} \\
$^1$SKL-MAIS, Institute of Automation, Chinese Academy of Sciences \\
$^2$School of Artificial Intelligence, University of Chinese Academy of Sciences \\
$^3$Beijing Academy of Artificial Intelligence, $^4$BeingBeyond \\
$^5$School of Computer Science, Peking University \\
\authorblockA{\authorrefmark{1} Correspondence: liboyu2021@ia.ac.cn, dongbin.zhao@ia.ac.cn}
% \thanks{This research was supported in part by NSFC under Grant 62450001, 62136008, Doubao Fund and the Suzhou Innovation and Entrepreneurship Leading Talents Programme - Innovation Leading Talent in Universities and Research Institutes with Grant No. ZXL2025310.}
}

%\author{\authorblockN{Michael Shell}
%\authorblockA{School of Electrical and\\Computer Engineering\\
%Georgia Institute of Technology\\
%Atlanta, Georgia 30332--0250\\
%Email: mshell@ece.gatech.edu}
%\and
%\authorblockA{Twentieth Century Fox\\
%\and
%\authorblockN{James Kirk\\ and Montgomery Scott}
%\authorblockA{Starfleet Academy\\
%San Francisco, California 96678-2391\\

% avoiding spaces at the end of the author lines is not a problem with
% conference papers because we don't use \thanks or \IEEEmembership

% \author{\authorblockN{Michael Shell\authorrefmark{1},
% Homer Simpson\authorrefmark{2},
% James Kirk\authorrefmark{3}, 
% Montgomery Scott\authorrefmark{3} and
% Eldon Tyrell\authorrefmark{4}}
% \authorblockA{\authorrefmark{1}School of Electrical and Computer Engineering\\
% Georgia Institute of Technology,
% Atlanta, Georgia 30332--0250\\ Email: mshell@ece.gatech.edu}
% \authorblockA{\authorrefmark{2}Twentieth Century Fox, Springfield, USA\\
% Email: homer@thesimpsons.com}
% \authorblockA{\authorrefmark{3}Starfleet Academy, San Francisco, California 96678-2391\\
% Telephone: (800) 555--1212, Fax: (888) 555--1212}
% \authorblockA{\authorrefmark{4}Tyrell Inc., 123 Replicant Street, Los Angeles, California 90210--4321}}

\maketitle

\begin{abstract}

Learning universal policies from cross-embodied data remains a fundamental challenge in robotics. Although Vision-Language-Action (VLA) models are pre-trained on large and diverse datasets, they typically rely on embodiment-specific fine-tuning to achieve strong performance in downstream tasks. This requirement severely limits their generalization capability and restricts knowledge transfer across embodiments performing similar tasks. To overcome these limitations, we focus on cross-embodied settings with shared robotic bases and heterogeneous end-effectors, and propose X-DiffVLA, a diffusion-based VLA model featuring a unified cross-embodied action head. X-DiffVLA can leverage the generative strengths of diffusion models to capture both the diversity and latent correlations in cross-embodied datasets. Specifically, we introduce Embodiment Forcing, a classifier-free guidance technique to implicitly steer action generation toward embodiment-specific functional components, capturing fine-grained structural nuances without explicit supervision. In addition, a Morphological Tree Diffusion approach is designed to strengthen behavioral correlations across diverse end-effectors, maximizing the transferability of heterogeneous demonstrations. Experimental results across RoboCasa and Isaac Gym, covering different embodiments from grippers to dexterous hands, show that X-DiffVLA achieves state-of-the-art performance, with improvements of 15.3\% and 12.5\%, respectively. Real-world evaluations further validate the robustness of the proposed framework and its effectiveness in scalable cross-embodied policy learning.
\end{abstract}

\IEEEpeerreviewmaketitle

\section{Introduction}

In recent years, training Vision-Language-Action (VLA) models on large-scale, heterogeneous, multi-task, and multi-embodiment datasets has emerged as a prominent research frontier~\cite{ghosh2024octo, xvla}. Models such as OpenVLA~\cite{kim2025openvla}, GR00T~\cite{gr00t}, and $\pi_{0.5}$~\cite{pi05} have demonstrated that large-scale pre-training equips models with superior visual understanding~\cite{chen2025see} and long-horizon task planning capabilities~\cite{li2025dualthor}, facilitating diverse robotic control tasks. However, existing architectures typically necessitate fine-tuning embodiment-specific action heads~\cite{gr00t} during downstream evaluation to ensure performance. Consequently, any variation in robot morphology requires the finetuning of a new action head. This paradigm not only impairs training efficiency, particularly in scenarios involving multiple end-effectors, but also obstructs cross-embodiment data utilization for similar tasks, leading to a landscape where specialized models excel in isolated tasks. Therefore, designing a cross-embodied action head for VLA is a unified path toward general intelligence in the field of robotics. 

\begin{figure}[!t]
    \centering %表示居中
    \includegraphics[width=\columnwidth]{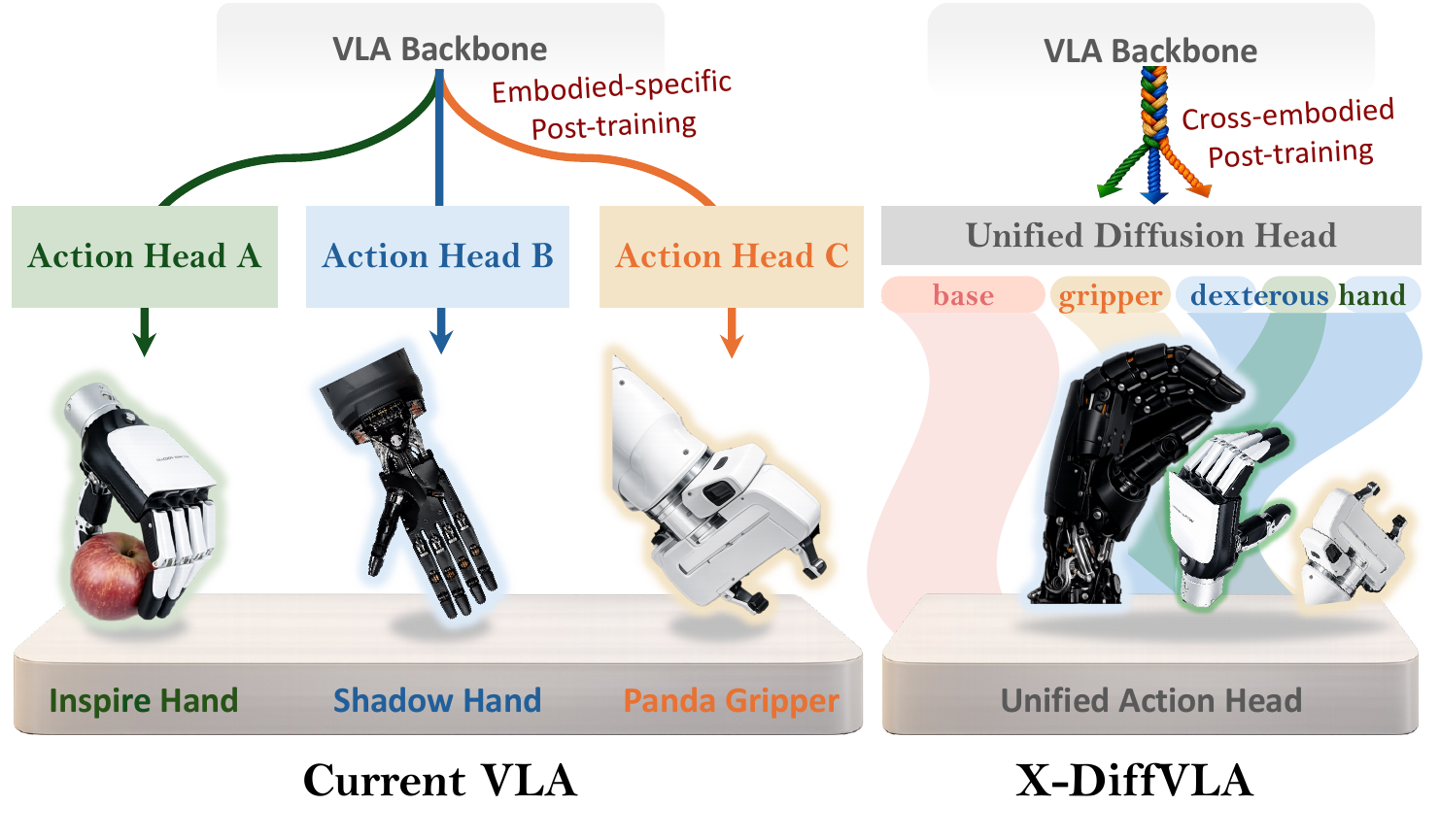}
    \vspace{-5mm}
    \caption{\textbf{The motivation of X-DiffVLA.} While embodied-specific post-training restricts VLA models to isolated tasks and end-effectors, our architecture enables cross-embodied learning and knowledge transfer, leveraging diverse data to transition from specialized experts to a unified, general robotic controller.}
    \label{overview}
    \vspace{-8mm}
\end{figure}

A fundamental challenge in developing a cross-embodied action head lies in effectively characterizing the difference across heterogeneous embodiments to facilitate morphology-aware action generation. Unlike visual~\cite{liu2025videos} or auditory modalities~\cite{shen2024naturalspeech}, robotic data exhibits substantial structural divergence~\cite{li2024mat}; for instance, the joint optimization across datasets including simple grippers and dexterous hands often precipitates distributional interference, leading to sub-optimal performance. Consequently, a control policy must possess morphological discrimination capabilities. \citet{RDT} demonstrates that utilizing a unified action space and diffusion transformer architectures is effective to leverage cross-gripper data on small-scale models. Extending to current VLA architecture with more end-effectors, achieving conditioned action generation for heterogeneous embodiments remains unexplored.
% Extending to VLA architecture, achieving conditioned action generation for more end-effectors is a critical challenge for maximizing data utility and enhancing policy generalization.

Effective knowledge transfer via cross-embodied action head is another challenge for VLA models. Cross-embodied data is valuable because there exist many tasks share underlying similarities across different embodiments~\cite{wei2024d}. A universal action head should learn shared representations from cross-embodiment data to enhance generalization. A representative scenario, frequently encountered in both simulation and real-world environments, involves a shared robotic platform equipped with diverse end-effectors~\cite{nasiriany2024robocasa, yakefu2025robochallenge}. Since the primary agents remain consistent, tasks across these embodiments exhibit significant functional similarities, suggesting an implicit correlation between their control policies. Prior research has demonstrated that this correlation exists and can be transferred; for instance, DemoGrasp~\cite{yuan2025demograsp} utilize motion retargeting to map control strategies from human hand poses to various dexterous hands. Similarly, \citet{bauer2025latent} find that learning a semantically aligned latent action space for dexterous hands, human hands, and parallel jaw grippers, can enable the optimization of a joint policy by refining only one single embodiment. Therefore, how cross-embodied action heads can effectively mine these latent correlations is another challenge we want to solve for achieving robust cross-embodied control within Vision-Language-Action (VLA) models.

% Diffusion models have demonstrated revolutionary performance in modeling multi-modal distributions, excelling in domains such as high-fidelity video generation and complex causal sequence inference. In robotics, the expressive power of diffusion allows models to capture the inherent stochasticity and multi-modal nature of human demonstrations. Nevertheless, existing diffusion-based policies are primarily applied to single-robot setups, limiting the transfer of physical interaction priors across different embodiments. This isolation prevents the exploitation of cross-embodied data, especially when dealing with varying action dimensions and temporal correlations. Consequently, adapting diffusion mechanisms to a generalized VLA framework requires a fundamental rethinking of how action sequences are decoupled and enforced across diverse morphologies.

To address these challenges, we present \textbf{X-DiffVLA}, a general VLA framework that enables cross-embodied post-training via a diffusion-based action head. By leveraging the capability of diffusion models to capture multi-peaked distributions in complex action spaces, X-DiffVLA can effectively accommodate diverse embodied data with a unified action space. We further propose \textbf{Embodied Forcing (EBF)}, a mechanism that integrates morphological information into the denoising process to guide action token generation. Additionally, a \textbf{Morphological Tree Diffusion (MPTD)} method is introduced to capture behavioral correlations across disparate end-effectors, facilitating cross-embodied knowledge transfer. Evaluated in both the task-complex RoboCasa and the high-dynamics Isaac Gym environments, spanning simple grippers to different dexterous hands, X-DiffVLA achieves state-of-the-art performance with improvements of 15.3\% and 12.5\%, respectively. Real-world experiments further underscore its potential as a scalable pathway toward universal robotic control.

In summary, our contributions encompass the following key advancements: \begin{enumerate}%[left=1em, itemsep=1pt] 
    \item We propose \textbf{X-DiffVLA}, a general VLA framework featuring a diffusion-based action head designed for post-training across heterogeneous robotic embodiments. 
    \item We introduce \textbf{Embodied Forcing} and \textbf{Morphological Tree Diffusion}, two key techniques that enhance both discrimination and behavioral association across diverse embodiments during the action generation process. 
    \item We conduct experiments in RoboCasa and Isaac Gym, demonstrating that X-DiffVLA significantly improves cross-embodied data utilization, achieving performance gains of 15.3\% and 12.5\% over state-of-the-art VLA baselines. 
\end{enumerate}

\section{Preliminaries}
\subsection{Action Heads in VLA Models}
% For current VLA models, action space can be categorized into two types: discrete action space and continuous action space~\cite{zhong2025survey}. The former discretizes the continuous action space based on predefined rules. A common strategy involves uniformly partitioning each dimension of the action space into 256 bins~\cite{kim2025openvla} and require the models to predict the probability distribution over these bins. However, since real action distributions are typically non-uniform, this discretization fails to provide fine-grained control signals. In contrast, mapping VLM features directly to a continuous action space via neural networks enables the generation of more diverse action signals.

% For continuous action space, three primary action heads are currently utilized: 

For current VLA models, continuous action space can be categorized into three types: the multilayer perceptron (MLP) head, the diffusion head, and the flow-matching head. The MLP head~\cite{liu2024robomamba} is straightforward but tends to learn the mean of various action distributions, that may be not allowed by physical executions. The diffusion head possesses superior capabilities for fitting multi-peaked distributions and has proven effective for robotic action generation in multi-task scenarios~\cite{ghosh2024octo, liu2025hybridvla}. The flow-matching head, adopted by current popular embodied models~\cite{pi05}, improves upon the diffusion head by directly predicting the denoising vector field, thereby achieving fewer sampling iterations and higher control frequencies.

The action head employed in X-DiffVLA is also improved from the diffusion head, leveraging its exceptional multi-modal fitting capabilities and conditional generation performance. We compare it with the flow-matching head in our experiments, demonstrating that the diffusion head exhibits a significant advantage under cross-embodied post-training conditions, as further substantiated in Section \ref{designchoice}.

\begin{figure*}[t]
    \centering %表示居中
    \includegraphics[width=0.9\textwidth]{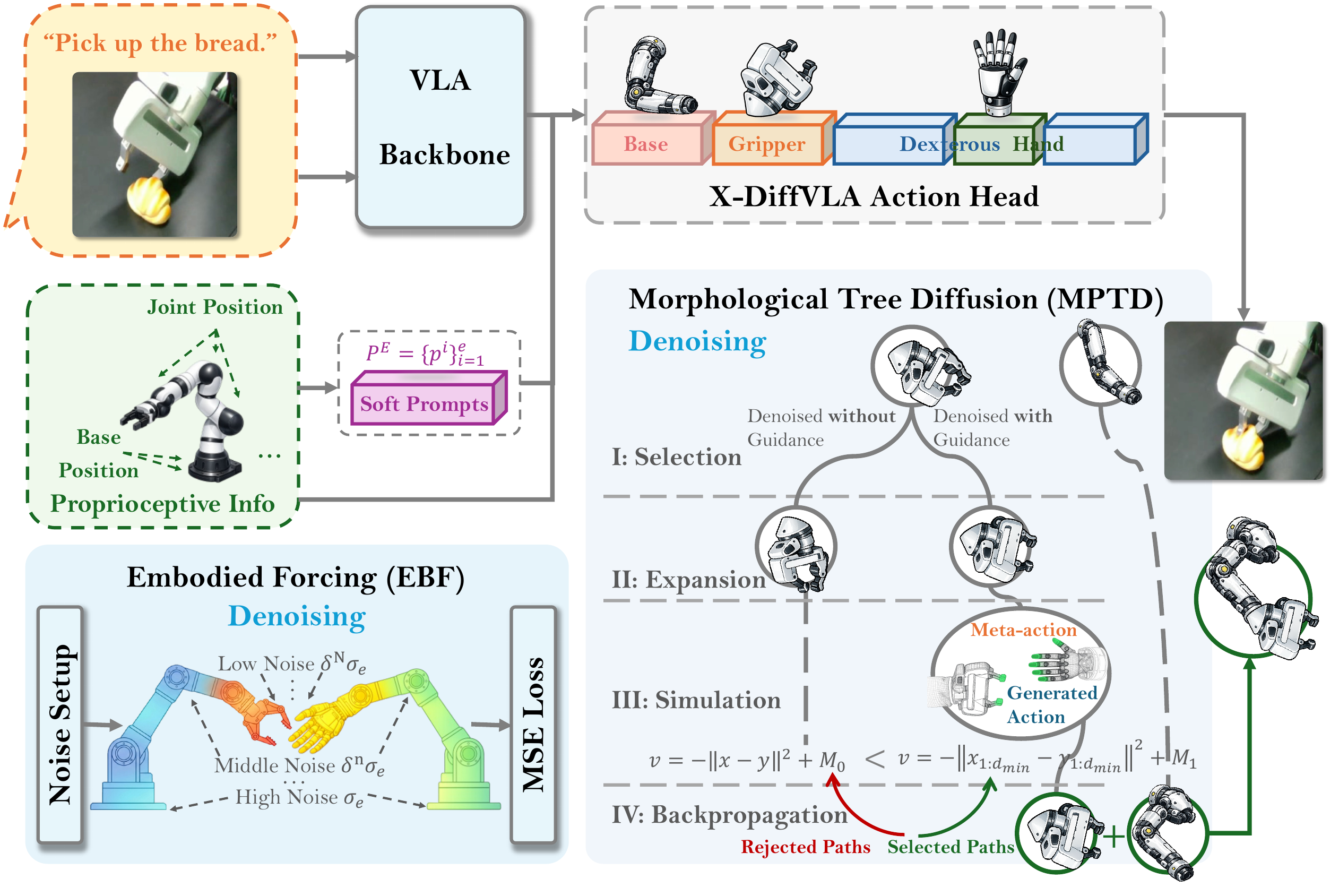}
    % \vspace{-3mm}
    \caption{\textbf{Architectural Overview of X-DiffVLA.} Our framework leverages a unified action space and a diffusion head to model multi-peaked action distributions for cross-embodied post-training. Key components include: (1) Embodied Forcing (EBF), which integrates morphological priors into the denoising process to enhance embodiment-specific discernment; and (2) Morphological Tree Diffusion (MPTD), designed to capture behavioral correlations across diverse end-effectors, thereby facilitating robust cross-embodied knowledge transfer.}
    \label{framework}
    \vspace{-6mm}
\end{figure*}

\subsection{Diffusion Generation for Multi-Peaked Distributions}
\label{diffusion}
Diffusion models have proven to be exceptional generative models for fitting multi-peaked data distributions~\cite{ho2020denoising}, while demonstrating robust conditional generation capabilities. The core optimization revolves around a denoising process. It first considers a forward diffusion process that systematically injects Gaussian noise into a data point over a sequence of timesteps. Let $q(\mathbf{x})$ represent the data distribution, where we define $\mathbf{x} \sim q$. This process is formalized as a Markov chain, where the data $\mathbf{x}$ at each step $k$ is incrementally added noise by:
\begin{equation}
q(\mathbf{x}^k | \mathbf{x}^{k-1}) = \mathcal{N}(\mathbf{x}^k; \sqrt{1 - \beta_k}\mathbf{x}^{k-1}, \beta_k \mathbf{I}),
\end{equation}
where $\mathcal{N}$ denotes the normal distribution and $\beta_k \in (0, 1)$ represents the noise variance at each step. This process continues until the original data is transformed into white noise at $\mathbf{x}^K$.

The reverse process is also a Markov chain, which aims to reconstruct the original data from noise using a parameterized model $p_\theta$:
\begin{equation}
p_\theta(\mathbf{x}^{k-1} | \mathbf{x}^k) = \mathcal{N}(\mathbf{x}^{k-1}; \boldsymbol{\mu}(\mathbf{x}^k, k), \gamma_k \mathbf{I}),
\end{equation}
where $\boldsymbol{\mu}$ is a neural network, and the covariance of it can be set to the identity scaled by a fixed constant $\gamma_k$. Furthermore, in the reverse transition $p_\theta(\mathbf{x}^{k-1} | \mathbf{x}^k)$, the model effectively directs the latent variables from a high-entropy Gaussian state toward the various local maximum modes of the underlying data manifold. Unlike GANs~\cite{goodfellow2020generative}, which often suffer from mode collapse by focusing on a single point estimate, the stochastic feature of the reverse paths ensures that the diffusion model can cover multiple high-probability regions of $q(\mathbf{x})$. This allows for the synthesis of diverse samples that faithfully represent the complex, multi-peaked distribution of the cross-embodied data.

\subsection{Monte Carlo Tree Diffusion}

Monte Carlo Tree Search (MCTS)~\cite{coulom2006efficient} is a planning algorithm widely used in Reinforcement Learning that combines tree search with stochastic simulations to effectively balance exploration and exploitation. Typically, MCTS consists of four stages: selection, expansion, simulation, and backpropagation. By iteratively exploring potential future states, MCTS allows an agent to estimate the value of long-term trajectories in complex decision-making environments. 

Since the diffusion denoising process is inherently stochastic, identical initial conditions sometimes yield significantly different outcomes. To enhance generation stability, Monte Carlo Tree Diffusion (MCTD)~\cite{yoonmonte} adapts the MCTS algorithm by employing the diffusion model as the core engine for state transitions and node expansion. During the search, the algorithm utilizes node-specific reward feedback to dynamically modulate the denoising guidance intensity. This mechanism deeply couples the generative capacity of diffusion models with the heuristic search of MCTS, ensuring thorough exploration of the search space while accelerating denoising convergence toward high-reward states. More related works can be seen in Appendix \ref{relatedwork}.

\section{Method}
The architectural framework of \textbf{X-DiffVLA} is illustrated in Fig.\ref{framework}. To facilitate effective cross-embodied post-training, we employ a unified action space and a diffusion head capable of modeling multi-peaked action distributions. We introduce Embodied Forcing to enhance the model’s ability to distinguish between diverse robotic embodiments and functional segments, ensuring precise conditional generation. This mechanism integrates specific morphological information directly into the denoising process to guide action token generation. Furthermore, to capture behavioral correlations across varied end-effectors, we propose Morphological Tree Diffusion, which promotes robust cross-embodied knowledge transfer in similar tasks.

\subsection{Embodied Forcing}
First, X-DiffVLA establishes a unified action space by defining a standardized representation that encompasses the maximum dimensions of robot bases, grippers, and dexterous hands. Within each category, various robot platforms share a common action subspace; for robots with lower degrees of freedom (DoF), the action vectors are aligned via zero-padding in the trailing dimensions. This partitioning of the unified action space is strategically designed to distinguish between locomotion and interaction tasks, thereby facilitating conditional generation by the diffusion head.

To address the challenge of convergence in multi-peaked optimization and achievement of morphology-guided action generation for cross-embodied data, X-DiffVLA introduces \textbf{Embodied Forcing (EBF)} technique. In the field of video generation, Diffusion Forcing~\cite{chen2024diffusion} is often employed to enhance the causality of diffusion models. While traditional diffusion models are trained using uniform noise levels across all tokens, Diffusion Forcing proposes training sequence diffusion models with independently varied noise levels per frame. This approach prioritizes the generation of significant inter-frame variations while minimizing fluctuations in static backgrounds. As a form of classifier-free guidance, it ultimately improves the ability of diffusion models to generate fine-grained local video details. Inspired by this, Embodied Forcing refines the initial noise distribution of the denoising process from the dual perspectives of global robotic structure and localized functional components.

\textbf{Global Structure-Aware Noise Initialization.}
Different embodied datas represent different distributions, therefore the diffusion head first need to realize the generation of robot structure-guided action. As we introduce in Section \ref{diffusion}, diffusion head generate actions through denosing process. Specifically, at each timestep $t$, the diffusion head predicts a $H$ steps action chunk $\tau_t$ conditioned on an embodiment feature $e$, which specifies the robot's morphology and dynamics:
\begin{equation}
    \tau_t = [a_t, a_{t+1}, \dots, a_{t+H}].
\end{equation}

We train the diffusion head, $\hat{\tau}_t = x_{0,\theta}(\tau_t^k, p_t, n_t, k, e)$, to reconstruct a clean trajectory, where $\tau_t^k$ is the original trajectory $\tau$ added with Gaussian noise at level $k$ in embodiment type $e$. $p_t$ represents the proprioceptive information, and $n_t$ represents the output tokens by VLM backbone. The forward process for a specific embodiment $e$ is defined as:
\begin{equation}
    \tau_t^k = \sqrt{1 - \beta_k}\tau_t^{k-1} + \sqrt{\beta_k} \cdot \sigma_e \epsilon,
    \epsilon \sim \mathcal{N}(0, I),
\end{equation}
where $\alpha_k$ is the hyper-parameter in traditional diffusion models and $\sigma_e$ represents the unique noise magnitude scale for embodiment $e$. This allows the diffusion head to characterize the difference across heterogeneous embodiments from different Gaussian noise regions.

\textbf{Local Function-Aware Noise Initialization.}
Beyond global structural differences across embodiments, the functional heterogeneity of robot components also leads to multi-peaked distributions in cross-embodied data. For example, end-effector action distribution demands higher precision than that of robot base. Therefore, we add function-guided noise injection to assist the diffusion head in generating more accurate, function-aligned actions.

We consider that the spatial causality of robot actions mirrors the temporal causality observed in videos. In video diffusion models, pyramid noise~\cite{chen2024diffusion} is typically applied to historical frames according to their temporal distance to diminish the influence of distant frames. Leveraging this intuition, we propose a spatial pyramid noise injection strategy, where noise scales are assigned based on a component’s proximity to the end-effector. The forward process is redefined as:
\begin{equation}
    \tau_t^k = \sqrt{1 - \beta_k}\tau_t^{k-1} + \sqrt{\beta_k} \cdot \boldsymbol{\Sigma}_{e,f} \epsilon, \epsilon \sim \mathcal{N}(0, I),  
\end{equation}

\begin{equation}
    \boldsymbol{\Sigma}_{e,f} =\text{diag}(\sigma_e, \delta\sigma_e, \delta^2\sigma_e, \dots), 
\end{equation}
where $\boldsymbol{\Sigma}_{e,f}$ represents a diagonal covariance matrix and $\sigma$ serves as the noise attenuation coefficient. In experiments, we find that a window-based pyramid outperforms a step-wise approach in conditional generation. This suggests that the underlying distribution of robotic action data is organized around functional blocks (e.g., arm, gripper, fingers) rather than simple spatial positions. Detailed experimental results are provided in Section \ref{parameter}.

Finally, the network is optimized using the Mean Squared Error (MSE) loss against the ground-truth clean trajectory:
\begin{equation}
    \mathcal{L} = \text{MSE}(x_{0,\theta}(\tau_t^k, p_t, n_t, k, e, f), \tau_t).
\end{equation}

\subsection{Morphological Tree Diffusion}

After solving the multi-peaked optimization problem of cross-embodied action heads, we further consider how to capture the correlation between different embodiments on similar tasks. This serves as the primary motivation for cross-embodied post-training to effectively mine transferable knowledge from heterogeneous robot data, thereby providing a robust solution to the distributional challenges in contemporary robotics datasets.

We propose \textbf{Morphological Tree Diffusion (MPTD)}, which extends the principles of MCTD to the processing of cross-embodied data. MCTD is a conditional diffusion generation framework designed for long-horizon planning tasks; it decomposes complex planning tasks into hierarchical sub-tasks and utilizes critical meta-points to guide the trajectory of the denoising path. By integrating MCTS and fast jumpy denoising techniques, MCTD efficiently gets the path values of sub-planning nodes, thereby steering the global denoising process toward higher-value plans. This approach effectively mitigates the inherent instability of diffusion denoising paths and provides a clue for mining action correlations across heterogeneous embodiments.

Under the influence of windowed pyramid noise via Embodied Forcing, the action space is functionally partitioned into several segments, which we define as the sub-tasks required for the diffusion-based generation process. To provide effective guidance, we construct meta-action nodes derived from the nearest-neighbor points of heterogeneous embodiments performing the same task. The value of these sub-tasks during the denoising process is evaluated based on the distance between joint positions. Given a sub-task generated action $\mathbf{x}$ and a reference meta-action $\mathbf{y}$, the node value $v$ is defined as follows:
% \begin{equation}
% \footnotesize
% v = 
% \begin{cases} 
% -\|\mathbf{x} - \mathbf{y}\|^2 + M_0, \text{if } \text{dim}(\mathbf{x}) = \text{dim}(\mathbf{y}) \\
% -\|\mathbf{x}_{1:d_{\min}} - \mathbf{y}_{1:d_{\min}}\|^2 + M_1, \text{if } \text{dim}(\mathbf{x}) \neq \text{dim}(\mathbf{y})
% \end{cases}
% \label{value}
% \end{equation}

\begin{equation}
\resizebox{0.9\columnwidth}{!}{ % 0.85表示缩放到行宽的85%
$v = 
\begin{cases} 
-\|\mathbf{x} - \mathbf{y}\|^2 + M_0, & \text{if } \text{dim}(\mathbf{x}) = \text{dim}(\mathbf{y}) \\
-\|\mathbf{x}_{1:d_{\min}} - \mathbf{y}_{1:d_{\min}}\|^2 + M_1, & \text{if } \text{dim}(\mathbf{x}) \neq \text{dim}(\mathbf{y})
\end{cases}$
}
\label{value}
\end{equation}
where $M_0$ and $M_1$ represent the base values of a node, with $M_1$ slightly exceeding $M_0$ to incentivize denoising conditioned on heterogeneous embodied data. $d_{\min}$ denotes the minimum functional dimensionality. This formulation is particularly critical for end-effector segments where zero-padding is utilized to account for morphological discrepancies. 

The denoising process is structured as an iterative tree search consisting of four distinct phases:

\begin{itemize}
   
    \item \textbf{Selection}: Starting from the white noise within the morphological tree, the diffusion head can either conduct self-denoising or leverage external embodiments as conditioning signals for the denoising process.

    \item \textbf{Expansion}: From the selected node, the diffusion model generates candidate action segments, conditioning on Embodied Forcing noise initialization.

    \item \textbf{Simulation}: To evaluate the expanded paths, we follow fast jumpy denoising to perform a rapid rollout evaluation. The potential value is caculated by Equation \ref{value}. 

    \item \textbf{Backpropagation}: The value derived from the simulation step is propagated back through the tree hierarchy.
\end{itemize}

MPTD ensures that the denoising gradient is strategically biased toward action distributions that maintain functional consistency across different robotic morphologies, effectively mining action correlations within cross-embodied datasets.

\subsection{Soft Prompts for Embodied Feature}

To enhance the discriminative capacity of X-DiffVLA across heterogeneous robots, we utilize a soft-prompting mechanism inspired by X-VLA framework~\cite{xvla}. Recognizing that diffusion-based models are particularly sensitive to the latent prompt space, we incorporate domain-specific learnable parameters $P^E = \{p^i\}_{i=1}^e$ to effectively absorb cross-embodiment heterogeneity. 
% Unlike rigid, template-based language prompts, these soft prompts are randomly initialized and implicitly optimized through end-to-end training to approximate a latent mapping of specific morphological configurations. By injecting these prompts at the early stages of action generation, we provide the diffusion backbone with fine-grained guidance, specifically bolstering the differentiation of diverse end-effectors and facilitating robust, cross-embodied learning.

\begin{table*}[h]
\caption{Results of X-DiffVLA in RoboCasa environment. We examine the impact of diffusion head under cross-embodied post-training by analyzing the reduction of two failure categories.}
\vspace{-1mm}
\centering
\begin{adjustbox}{width=0.9\linewidth, center} % Slightly increased width to accommodate new columns
\renewcommand{\arraystretch}{0.9}
\begin{tabular}{l ccc c cc}
\toprule
\multirow{2}{*}{\textbf{Method}} & \multicolumn{3}{c}{\textbf{Success Rate}} & \multirow{2}{*}{\textbf{Avg.}} & \multicolumn{2}{c}{\textbf{Failure Categories}} \\
\cmidrule(lr){2-4} \cmidrule(lr){6-7}
& \textbf{Panda} & \textbf{Robotiq-85} & \textbf{Inspire} & & \#1 & \#2 \\
\midrule
UniVLA          & 39.5\% & 37.0\% & 35.1\% & 37.2\% & 889 & 241 \\
XR-1           & 45.2\% & 43.8\% & 34.9\% & 41.3\% & 863 & 193 \\
X-VLA           & 50.1\% & 46.9\% & 45.8\% & 47.6\% & 816 & \cellcolor{tableblue}{127} \\
GR00T-N1        & 39.6\% & 40.4\% & 38.5\% & 39.5\% & 854 & 235 \\
$\pi_0$ +FAST         & 44.3\% & 45.8\% & 39.2\% & 43.1\% & 820 & 204 \\
$\pi_0$         & 45.1\% & 45.2\% & 37.8\% & 42.7\% & 834 & 197 \\
$\pi_{0.5}$    & 52.3\% & 51.1\% & 44.2\% & 49.2\% & \cellcolor{tablepeach}{784} & 130 \\
X-DiffVLA(Ours) & \textbf{67.2\%} & \textbf{68.0\%} & \textbf{58.3\%} & \textbf{64.5\%} & \cellcolor{tablepeach}{550} & \cellcolor{tableblue}{89} \\
\bottomrule
\end{tabular}
\end{adjustbox}
\label{robocasa}
\vspace{-6mm}
\end{table*}

\section{Experiments}
\subsection{Experiments Setup}
To evaluate the generalizability of our method across diverse robotic platforms, we conduct experiments in Robocasa~\cite{nasiriany2024robocasa}, Isaac Gym~\cite{makoviychuk2021isaac}, and real-world environments. While Robocasa serves as a high-fidelity benchmark with rich task scenarios and object rendering, its native support is limited to two parallel grippers and one dexterous hand. To increase the comprehensiveness of the experimental results, we integrate Isaac Gym to expand our evaluation of X-DiffVLA to a broader range of embodiments, including one gripper and two distinct types of dexterous hands across ten unique objects. These results demonstrate that X-DiffVLA exhibits robust universality when facing both cross-embodiment diversity and task complexity, ensuring a comprehensive evaluation of the model. Finally, we deploy X-DiffVLA on both parallel-gripper and dexterous-hand robots in real-world environments. This deployment serves to validate the effectiveness of our model's transition from simulation to physical hardware.

We utilize Being-H0 2B~\cite{luo2025being} as the backbone model to validate the performance of our action head. Within the Robocasa benchmark, we employ the Rethink Mount and GR1 as robotic bases, paired with Panda gripper, Robotiq-85 gripper, and Inspire hands as end-effectors. For Isaac Gym, we follow the experimental configuration of DexGraspNet~\cite{zhang2024dexgraspnet}, utilizing Panda gripper, Inspire hands, and Shadow hands to perform grasping tasks across ten distinct objects. Finally, for real-world validation, we deploy the Panda gripper and Inspire hands mounted on a 7-DoF Franka Research 3 (FR3) arm.

\subsection{Baselines}
To rigorously evaluate the performance of \textbf{X-DiffVLA}, we compare it with several baselines that represent different paradigms in VLA models:
\begin{itemize}
    \item \textbf{GR00T-N1~\cite{gr00t}, $\pi_0$~\cite{pi_0}, $\pi_0$+FAST, $\pi_{0.5}$~\cite{pi05}} represent the state-of-the-art (SOTA) methods in this field, employing flow-matching action heads. 
    \item \textbf{UniVLA~\cite{univla}, X-VLA~\cite{xvla}, XR-1~\cite{fan2025xr}} constitute the mainstream research in cross-embodied VLA. These frameworks facilitate cross-embodied knowledge transfer in pre-train stage through latent action representations, soft-prompt-based hardware descriptions, and unified video-motion encoding, respectively.
\end{itemize}
To maintain evaluation fairness, all models are post-trained using a unified action space across heterogeneous embodiments. Detailed configurations are available in the Appendix \ref{setup details}.

\subsection{Experimental Results}

\subsubsection{\textbf{Is Diffusion Head Effective to Handle Multi-peaked Distributions in Different Embodiments}} As a generative model capable of fitting multimodal distributions, diffusion models provide the theoretical foundation for generating action distributions tailored to diverse robotic embodiments. Within the Robocasa environment, we establish a benchmark based on the GR00T dataset, encompassing 30 task categories for each end-effector, as detailed in Appendix \ref{setup details}. For each task, we perform post-training using a mixture of 50 trajectories. Evaluation is conducted through 20 test trials per task, with the results illustrated in the TABLE \ref{robocasa}.

Experimental results demonstrate that X-DiffVLA achieves superior performance, validating the effectiveness of diffusion models in modeling cross-embodiment distributions. To further analyze the diffusion head, we conduct an error analysis on failed task trajectories, categorizing them into two failure types: Mobility Failure (Failure \#1) and Interaction Failure (Failure \#2). Specifically, Failure \#1 occurs when the end-effector fails to reach an approachable range of the object, whereas Failure \#2 involves operational errors despite the end-effector being within the interaction workspace. The results indicate that X-DiffVLA significantly reduces the probability of both failure types, particularly Mobility Failures. This underscores its ability to achieve functional partitioning of action space and facilitates the global optimization of multi-peaked distributions. What's more, this provides empirical evidence that X-DiffVLA facilitates positive knowledge transfer across embodiments, whereby post-training on heterogeneous data significantly augments the model's proficiency in executing task-specific behaviors for a given robot morphology.

\subsubsection{\textbf{What enables Diffusion Models to handle diverse embodiment distributions}}
\label{parameter}
In experiments, we find that EBF is critical for X-DiffVLA, and it exhibits extreme sensitivity to noise parameters. We find that the noise technique acts as a latent guide, enabling the diffusion head to fit cross-embodiment distributions and facilitate conditional generation. We conduct a series of sensitivity tests on the pyramid noise hyperparameters, the results are summarized in the TABLE \ref{hyperparameter_analysis}. It shows that while excessively large decay coefficients reduce noise discriminability and hinder functional partitioning, overly small coefficients weaken causal coherence in the action space, resulting in unexecutable action generation. Furthermore, the impact of window sizes suggests that the diffusion head favors a functional decomposition of action generation, providing a robust empirical basis for task partitioning in MPTD. A window size of $W_n=3$ partitions the action space into [base, gripper, hand], whereas $W_n=9$ denotes a full decomposition of base by degrees of freedom; notably, neither achieves the best results.

\vspace{-1mm}
\begin{table}[h]
\centering
\caption{Performance comparison under different hyperparameter settings ($\delta$ and $W_n$) in RoboCasa environment. When evaluating one variable, the other is held at its optimal value to ensure the reliability of the experimental results.}
\vspace{-1mm}
\begin{adjustbox}{width=\columnwidth, center}
\begin{tabular}{c cccc}
\toprule
\multirow{2}{*}{\textbf{Hyperparameter}} & \multicolumn{3}{c}{\textbf{Success Rate}} & \multirow{2}{*}{\textbf{Avg.}} \\
\cmidrule(lr){2-4}
& \textbf{Panda} & \textbf{Robotiq-85} & \textbf{Inspire} & \\
% \textbf{Hyperparameter} & \textbf{Panda} & \textbf{Robotiq-85} & \textbf{Inspire} & \textbf{Avg.} \\
\midrule
$\delta=0.9$           & 59.5\% & 61.7\% & 45.9\% & 55.7\% \\
$\delta=0.5$           & \textbf{67.2\%} & \textbf{68.0\%} & \textbf{58.3\%} & \textbf{64.5\%}\\
$\delta=0.1$           & 41.3\% & 39.5\% & 31.7\% & 37.5\% \\
\midrule
\addlinespace
$W_n=9$  & 43.6\% & 46.7\% & 41.1\% & 43.8\% \\
$W_n=4$  & \textbf{67.2\%} & \textbf{68.0\%} & \textbf{58.3\%} & \textbf{64.5\%}\\
$W_n=3$  & 46.3\% & 45.5\% & 37.2\% & 43.0\% \\
\bottomrule
\end{tabular}
\end{adjustbox}
\label{hyperparameter_analysis}
\vspace{-3mm}
\end{table}

\subsubsection{\textbf{Are EBF Useful for Flow-Matching Action Head}}
\label{designchoice}
Given that flow-matching heads are diffusion-based and increasingly adopted in VLA architectures, we investigate whether noise-guided generation methods can yield similar performance gains. To this end, we conduct parallel experiments within the RoboCasa environment. Specifically, EBF is applied during the noise initialization phase of the flow-matching head, with results summarized in Table \ref{flow-matching}. Our findings indicate that EBF fails to effectively enhance cross-embodiment generalization for flow-matching architecture. A plausible explanation is that the denoising process in diffusion models may need a more granular sampling during the high-noise initial stages, which are critical for capturing robot-specific structural constraints and generating corresponding actions. Conversely, although flow-matching also uses the EBF method, its primary advantage lies in efficient few-step sampling. In cross-embodied tasks, the simplified probability paths of flow-matching may struggle to capture the structural and functional variances across different robotic morphologies.

\begin{table}[h]
\caption{Study of EBF under diffusion head (DH) and flow-matching head (FH). For all configurations, a unified action space is employed, and the MPTD approach is omitted.}
\vspace{-1mm}
\centering
\begin{adjustbox}{width=\columnwidth, center} 
\begin{tabular}{l ccc c}
\toprule
\multirow{2}{*}{\textbf{Method}} & \multicolumn{3}{c}{\textbf{Success Rate}} & \multirow{2}{*}{\textbf{Avg.}} \\
\cmidrule(lr){2-4}
& \textbf{Panda} & \textbf{Robotiq-85} & \textbf{Inspire} & \\
\midrule
FH & 43.3\% & 43.1\% & 38.1\% & 41.5\% \\
FH + EBF & 42.5\% & 44.7\% & 40.6\% & 42.6\% \\
DH & 41.2\% & 39.4\% & 41.8\% & 40.8\% \\
DH + EBF & \textbf{56.9\%} & \textbf{58.3\%} & \textbf{47.1\%} & \textbf{54.1\%} \\
\bottomrule
\end{tabular}
\end{adjustbox}
\label{flow-matching}
\vspace{-2mm}
\end{table}

\subsubsection{\textbf{What's the Relationship Between Different Embodiments Data}}
To further enhance the versatility of the diffusion head across more end-effectors, we conduct additional evaluations within the Isaac Gym environment using a diverse set of robotic hands, including the Panda gripper, Inspire hand, and Shadow hand. Following the benchmarking and data sourcing protocols established in DemoGrasp~\cite{yuan2025demograsp}, we select 10 distinct objects, with 10 trajectories per object utilized for model post-training. More details can be seen in Appendix \ref{setup details}. These experiments are executed to investigate whether X-DiffVLA can identify latent correlations between the actions of low-dimensional grippers and high-dimensional dexterous hands, thereby improving both model performance and generalization. The results are illustrated in the TABLE \ref{issacgym}.

\begin{table}[h]
\caption{Results of X-DiffVLA in Isaac Gym environment. Additional dexterous hand data is incorporated to evaluate the model's capability in capturing correlations across heterogeneous embodiments.}
\vspace{-1mm}
\centering
\begin{adjustbox}{width=\columnwidth, center} 
\renewcommand{\arraystretch}{1.2}
\begin{tabular}{l ccc c}
\toprule
\multirow{2}{*}{\textbf{Method}} & \multicolumn{3}{c}{\textbf{Success Rate}} & \multirow{2}{*}{\textbf{Avg.}} \\
\cmidrule(lr){2-4}
& \textbf{Panda} & \textbf{Inspire} & \textbf{Shadow} & \\
\midrule
X-VLA           & 56.5\% & 57.0\% & 58.0\% & 57.2\% \\
$\pi_0$         & 55.5\% & 52.5\% & 54.0\% & 54.0\% \\
$\pi_{0.5}$    & 59.5\% & 58.5\% & 57.5\%  & 58.5\% \\
\textbf{X-DiffVLA (Ours)} & \textbf{73.5\%} & \textbf{69.5\%} & \textbf{70.0\%} & \textbf{71.0\%} \\
\bottomrule
\end{tabular}
\end{adjustbox}
\label{issacgym}
\vspace{-3mm}
\end{table}

\begin{figure*}[t]
  \centering
  \begin{minipage}[b]{0.32\textwidth}
    \centering
    \includegraphics[width=\textwidth]{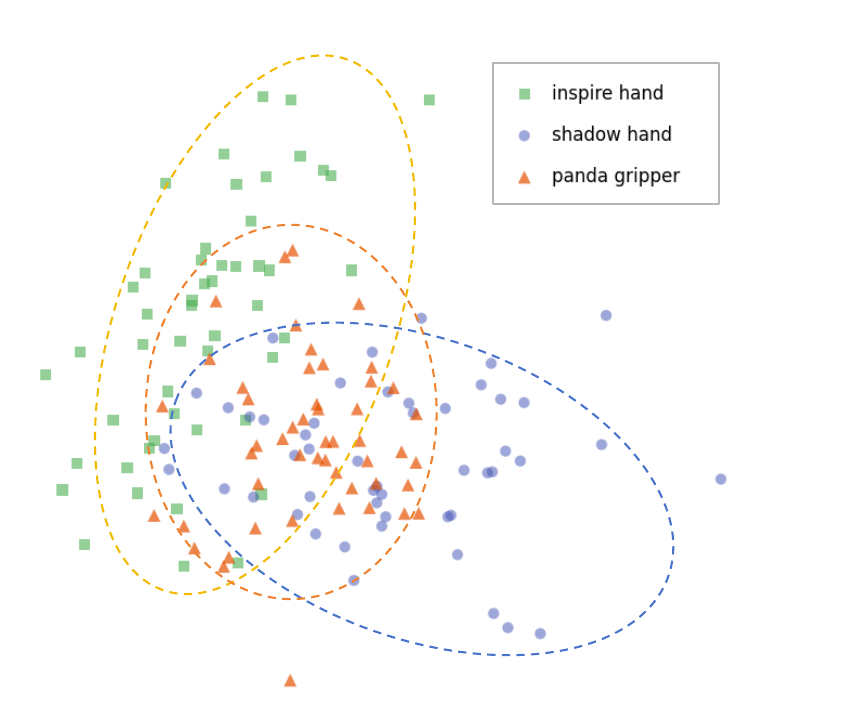}
    \caption{\textbf{T-SNE visualization results of three embodiments in Isaac Gym.} The visualization compares the final joint positions of different end-effectors over 50 trials.}
    \label{action}
  \end{minipage}
  \hfill 
  \begin{minipage}[b]{0.65\textwidth}
    \centering
    \includegraphics[width=\textwidth]{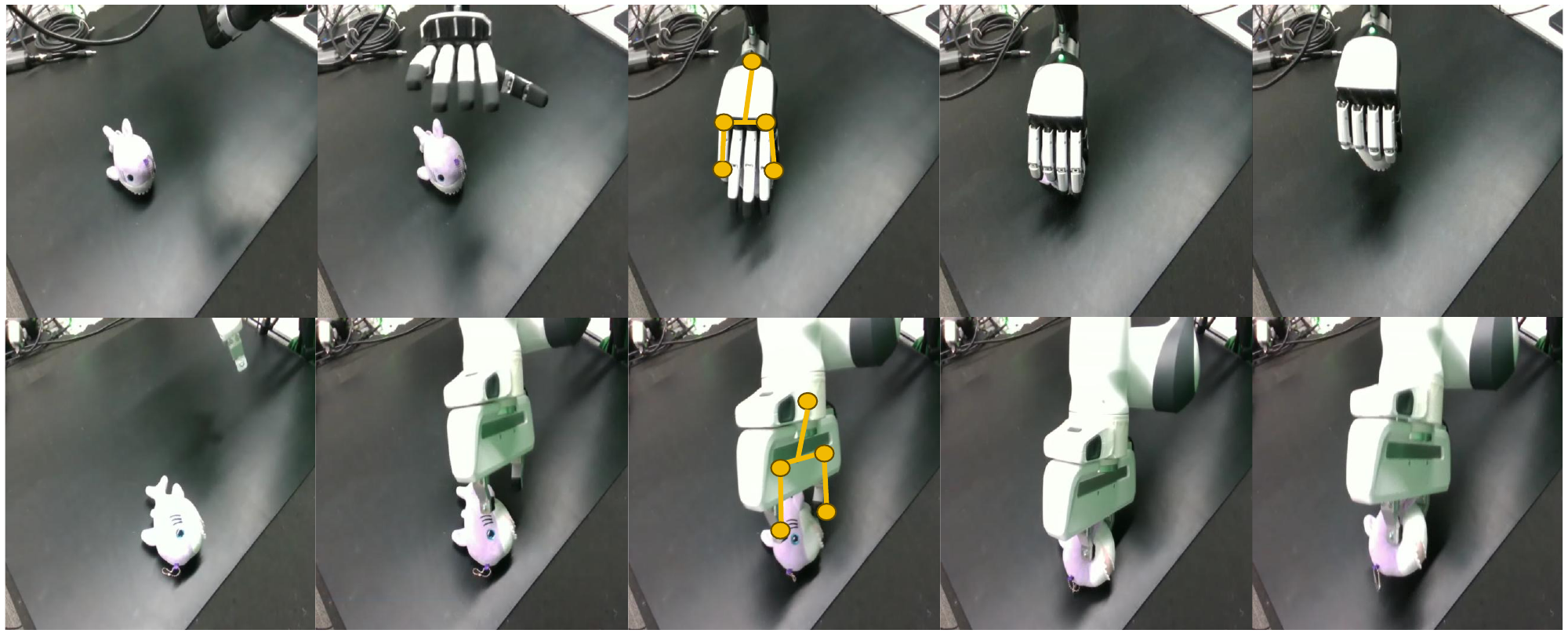}
    \caption{\textbf{Visualization results of real world experiments.} The Inspire hand demonstrates a policy learned from parallel gripper data (highlighted in yellow). Such behavior underscores the effectiveness of X-DiffVLA in leveraging cross-embodied datasets to facilitate robust knowledge transfer.}
    \label{realrobot}
  \end{minipage}
  \vspace{-5mm}
\end{figure*}

To provide a clearer analysis of the action generation results across different embodiments, we employ T-SNE~\cite{maaten2008visualizing} to visualize the final action distributions, as illustrated in Fig.\ref{action}. To ensure consistency with the gripper’s positions, we select the joint positions of the two most distant fingertips as the primary features for dexterous hands. Each task is evaluated over 50 trials, with the final joint positions being recorded for analysis, and we only choose one of them present in the paper. We can observe that despite the significant structural disparities between parallel grippers and dexterous hands, their action spaces exhibit a high degree of correlation. This demonstrates that X-DiffVLA successfully facilitates cross-embodiment knowledge transfer from similar tasks with the effective guidance of MPTD.

\subsection{Real Robots}
To evaluate the effectiveness of the X-DiffVLA action head, we conduct real-world validation using both Panda grippers and Inspire hands. For our hardware platform, we employ a unified FR3 robotic arm as the primary base. Real-robot datasets are collected via teleoperation, utilizing a GELLO device for arm control and Manus gloves for hand manipulation. More details can be seen in Appendix \ref{setup details}. For post-training, we collect 10 trajectories for each of the 5 distinct objects in the pick-up tasks. And each task is rigorously evaluated over 20 independent trials. The experimental results are presented in the TABLE \ref{real} and Fig.\ref{realrobot}.

\begin{table}[h]
\caption{Results of X-DiffVLA in real world experiments.}
\vspace{-1mm}
\centering
\begin{adjustbox}{width=0.95\linewidth, center} 
\renewcommand{\arraystretch}{1.0}
\begin{tabular}{l cc c}
\toprule
\multirow{2}{*}{\textbf{Method}} & \multicolumn{2}{c}{\textbf{Success Rate}} & \multirow{2}{*}{\textbf{Avg.}} \\
\cmidrule(lr){2-3}
& \textbf{Panda} & \textbf{Inspire} & \\
\midrule
X-VLA           & 51.0\% & 47.0\% & 49.0\% \\
$\pi_0$         & 55.0\% & 47.0\% & 51.0\% \\
$\pi_{0.5}$   & 59.0\% & 55.0\% & 57.0\% \\
\textbf{X-DiffVLA (Ours)} & \textbf{67.0\%} & \textbf{60.0\%} & \textbf{63.5\%} \\
\bottomrule
\end{tabular}
\end{adjustbox}
\label{real}
\vspace{-2mm}
\end{table}

From the Fig.\ref{realrobot}, we can observe that the Inspire hand prioritizes stabilizing objects using its two most distant fingers (the yellow part in the third image) before engaging the remaining digits for a full grasp. This behavior closely mirrors the grasping strategy characteristic of parallel grippers, providing compelling evidence that X-DiffVLA effectively leverages cross-embodiment data. Such strategic alignment further demonstrates the model's capacity for robust knowledge transfer between different robotic morphologies.

\subsection{Ablation Study}

We conduct ablation studies in the RoboCasa environment to investigate the individual contributions of each proposed module to X-DiffVLA. Specifically, w/o SP denotes the removal of the Soft Prompt mechanism; w/o MPTD and w/o EBF represent the models excluding Morphological Tree Diffusion and Embodied Forcing, respectively; and w/o ALL serves as the baseline utilizing only the initial diffusion head.

As presented in Table \ref{ablatioin}, the results demonstrate that both EBF and MPTD are indispensable, as they effectively capture the inherent disparities and underlying correlations across different embodiments, respectively. Furthermore, the soft prompt provides effective guidance for conditional generation, enabling the model to generate more precise actions tailored to specific robotic morphologies.

\begin{table}[h]
\caption{Ablation study of X-DiffVLA components in the RoboCasa environment. The experimental setup is consistent with the main experiments.}
\centering
\begin{adjustbox}{width=\linewidth, center} 
\renewcommand{\arraystretch}{1.1}
\begin{tabular}{l ccc c}
\toprule
\multirow{2}{*}{\textbf{Method}} & \multicolumn{3}{c}{\textbf{Success Rate}} & \multirow{2}{*}{\textbf{Avg.}} \\
\cmidrule(lr){2-4}
& \textbf{Panda} & \textbf{Robotiq-85} & \textbf{Inspire} & \\
\midrule
X-DiffVLA & \textbf{67.2\%} & \textbf{68.0\%} & \textbf{58.3\%} & \textbf{64.5\%}\\
w/o SP & 66.1\% & 65.4\% & 55.4\% & 62.3\% \\
w/o MPTD & 56.9\% & 58.3\% & 47.1\% & 54.1\% \\
w/o EBF & 46.7\% & 48.3\% & 35.5\% & 43.5\% \\
w/o ALL & 41.2\% & 39.4\% & 41.8\% & 40.8\% \\
\bottomrule
\end{tabular}
\end{adjustbox}
\label{ablatioin}
\end{table}

\section{Conclusion}
In this paper, we introduce X-DiffVLA, a novel and general VLA framework designed to support cross-embodied post-training through a unified diffusion-based action head. By integrating the Embodied Forcing (EBF) mechanism and Morphological Tree Diffusion (MPTD), our approach successfully captures both the unique morphological discrimination and the underlying behavioral correlations to generate more precise actions. Extensive evaluations in RoboCasa and Isaac Gym demonstrate that X-DiffVLA significantly outperforms existing baselines, achieving an average improvement of 15.3\% and 12.5\%, respectively. Furthermore, our real-world experiments validate the model's ability to facilitate effective knowledge transfer between grippers and dexterous hands. These results suggest that X-DiffVLA provides a scalable and robust pathway toward universal robotic control, laying the groundwork for more versatile and generalized embodied AI systems.

\section{Acknowledgement}
This work was supported by NSFC in part under Grant 62136008, 62450001 and 62476008. This work was also supported by Beijing Major Science and Technology Project under Contract no.Z251100008125023 and Beijing Academy of Artificial Intelligence(BAAI).

%% Use plainnat to work nicely with natbib. 

\bibliographystyle{plainnat}
\bibliography{paper_template}

\cleardoublepage
\appendix
\subsection{Experiments Setup}
\label{setup details}
\subsubsection{\textbf{RoboCasa}}
RoboCasa~\cite{nasiriany2024robocasa} is a large-scale simulation framework for training generally capable robots to perform everyday tasks. It features realistic and diverse human-centered environments with a focus on kitchen scenes. It provides over 2500 3D assets across 150+ object categories and dozens of interactable furniture and appliances. The simulator supports many end-effectors of shared base, such as Panda, UR5e, Sawyer and so on. 

Based on the GR00T~\cite{gr00t} dataset, we select two distinct robot bases, RethinkMount and GR1ArmsOnly, each configured with three different types of end-effectors: the Robotiq85Gripper, the PandaGripper, and Inspire-Hands. To conduct validation experiments, we design a suite of 30 tasks for each end-effector configuration, as detailed in the TABLE \ref{tab:task_list}. Our data generation pipeline involves performing trajectory playback on the original dataset, followed by motion retargeting to map these trajectories onto the new end-effectors. To ensure the integrity of the processed data, every trajectory is verified through environmental feedback and human selection to guarantee successful task completion and collision-free execution.

For the post-training process, we collect 50 trajectories for each task and end-effector combination. The models are trained using the Being-H0 2B backbone~\cite{luo2025being}. The specific hyper-parameters used are detailed in TABLE \ref{tab:hyperparams}.

\begin{table}[!h]
\centering
\renewcommand{\arraystretch}{1.1}
\caption{Hyper-parameters for post-training experiments in RoboCasa environment.}
\vspace{-2mm}
\label{tab:hyperparams}
\small
\begin{tabular}{lc}
\toprule
\textbf{Parameter} & \textbf{Value} \\ 
\midrule
Input Image Size & $224 \times 224$ \\
Training Iterations & 50k \\
Batch Size & 32 \\
Learning Rate & $2 \times 10^{-5}$ \\
Weight Decay & 0.05 \\
Optimizer & AdamW \\
Precision & BF16 \\
Diffusion Model Objective & $x_0$-prediction \\
Selection Scale & 3 \\ 
Noise Schedule & Cosine \\
Noise decay & 0.5 \\
Window numbers & 4 \\
Sampling steps & 50 \\
Jumpy Denoising Interval & 10 \\
\bottomrule
\end{tabular}
\end{table}

\begin{table}[t]
\centering
\renewcommand{\arraystretch}{1.1}
\caption{Task list of 30 validation tasks for each end-effector in RoboCasa environment.}
\vspace{-2mm}
\label{tab:task_list}
\small
\begin{tabular}{ll}
\toprule
\textbf{Robot Base} & \textbf{Task Category} \\ 
\midrule
\multirow{12}{*}{RethinkMount} & CloseDoubleDoor \\
 & CloseDrawer \\
 & CloseSingleDoor \\
 & CoffeePressButton \\
 & CoffeeServeMug \\
 & OpenDoubleDoor \\
 & OpenDrawer \\
 & OpenSingleDoor \\
 & PnPFromCabToCounter \\
 & PnPFromCounterToCab \\
 & PnPFromCounterToStove \\
 & PnPFromStoveToCounter \\ 
\midrule
\multirow{18}{*}{GR1ArmsOnly} & PnPBottleToCabinet \\
 & PnPCanToDrawer \\
 & PnPCupToDrawer \\
 & PnPPotatoToMicrowave \\
 & PnPWineToCabinetClose\\
 & PnPFromCounterToMicrowave \\
 & PnPFromCounterToSink \\
 & PnPFromCuttingboardToBasket \\
 & PnPFromCuttingboardToCardboardbox \\
 & PnPFromCuttingboardToPan \\
 & PnPFromCuttingboardToPot \\
 & PnPFromCuttingboardToTieredbasket \\
 & PnPFromMicrowaveToCounter \\
 & PnPFromPlacematToBasket \\
 & PnPFromPlacematToBowl \\
 & PnPFromPlacematToPlate \\
 & PnPFromPlateToPan \\
 & PnPFromSinkToCounter \\ 
\bottomrule
\end{tabular}
\end{table}

\subsubsection{Issac Gym}
To further evaluate the generalization capabilities of X-DiffVLA, we introduce an additional experimental environment based on Isaac Gym. This setup aims to demonstrate that X-DiffVLA can perform cross-embodied post-training across a broader range of complex dexterous hand structures. Following the experimental protocol established by DemoGrasp~\cite{yuan2025demograsp}, we select pick-up tasks involving 10 distinct object categories. 

The hardware configuration consists of a Franka Research 3 (FR3) robot arm  as the base, integrated with three different end-effectors: the Panda gripper, Inspire hands, and Shadow hands. For each object category, we leveraged the DemoGrasp policy to collect 10 expert trajectories with each object for post-training. The specific training configurations and hyper-parameters are shown in the TABLE \ref{tab:hyperparams-issac}.

\begin{table}[!h]
\centering
\renewcommand{\arraystretch}{1.1}
\caption{Hyper-parameters for post-training experiments in Isaac Gym environment.}
\vspace{-2mm}
\label{tab:hyperparams-issac}
\small
\begin{tabular}{lc}
\toprule
\textbf{Parameter} & \textbf{Value} \\ 
\midrule
Input Image Size & $224 \times 224$ \\
Training Iterations & 50k \\
Batch Size & 32 \\
Learning Rate & $5 \times 10^{-5}$ \\
Weight Decay & 0.05 \\
Optimizer & AdamW \\
Precision & BF16 \\
Diffusion Model Objective & $x_0$-prediction \\
Selection Scale & 3 \\ 
Noise Schedule & Cosine \\
Noise decay & 0.6 \\
Window numbers & 6 \\
Sampling steps & 50 \\
Jumpy Denoising Interval & 10 \\
\bottomrule
\end{tabular}
\end{table}

\subsubsection{Real World}
\begin{figure}[t]
  \centering
  % 第一行
  \begin{subfigure}[b]{0.48\columnwidth}
    \centering
    \includegraphics[width=\textwidth]{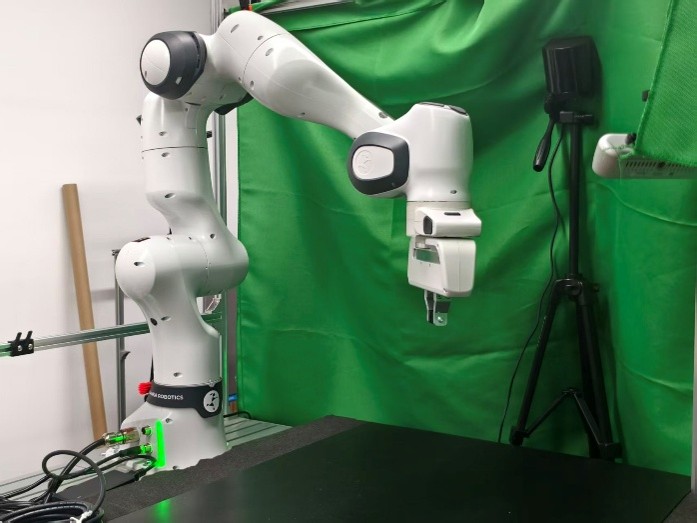}
    \caption{Overview of real-world experiments setup.}
    \label{fig:real_base}
  \end{subfigure}
  \hfill
  \begin{subfigure}[b]{0.48\columnwidth}
    \centering
    \includegraphics[width=\textwidth]{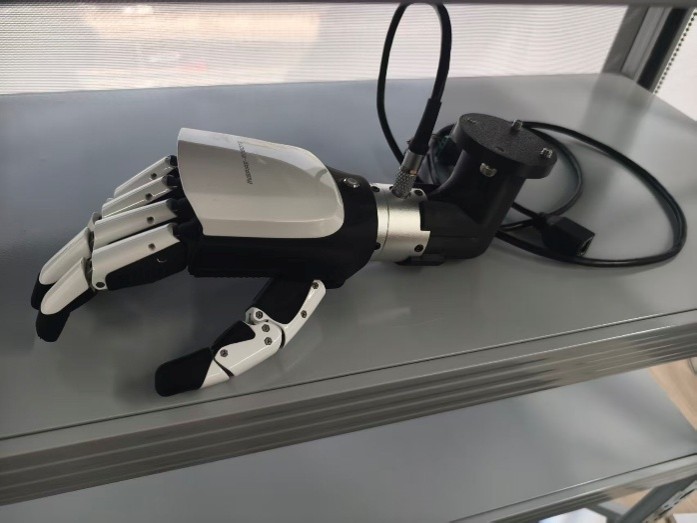}
    \caption{Visualization of inspire hands.}
    \label{fig:real_inspire}
  \end{subfigure}
  \vspace{20pt} % 上下行之间的间距
  % 第二行
  \begin{subfigure}[b]{0.48\columnwidth}
    \centering
    \includegraphics[width=\textwidth]{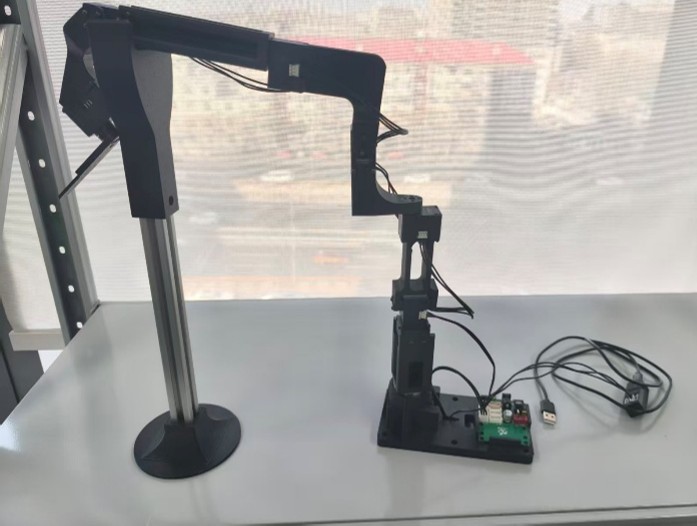}
    \caption{Visualization of the Gello arm.}
    \label{fig:gello}
  \end{subfigure}
  \hfill
  \begin{subfigure}[b]{0.48\columnwidth}
    \centering
    \includegraphics[width=\textwidth]{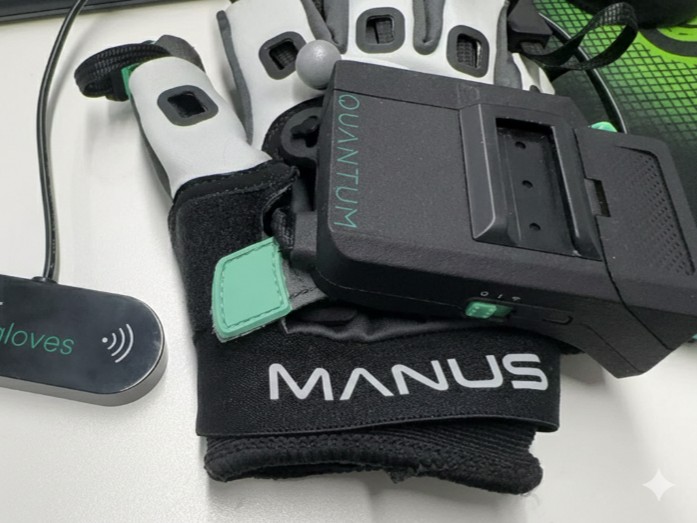}
    \caption{Visualization of Manus gloves.}
    \label{fig:manus}
  \end{subfigure}
  \vspace{-5mm}
  \caption{\textbf{Real-world experimental platforms and hardware for data collection.}}
  \label{fig:overall_setup}
\end{figure}

To evaluate the effectiveness of the X-DiffVLA action head, we conduct real-world validation using both Panda grippers and Inspire hands mounted on a FR3 robotic arm, as shown in Fig.\ref{fig:overall_setup}. Our real-robot datasets are collected via teleoperation, employing a GELLO device for arm control and Manus gloves for dexterous hand manipulation.

% \begin{figure}[t]
%   \centering
%   \begin{minipage}[b]{0.49\columnwidth}
%     \centering
%     \includegraphics[width=\textwidth]{img/real_base.jpg}
%     \caption{\textbf{Overview of real-world experiments setup.}}
%     \label{action}
%   \end{minipage}
%   \hfill 
%   \begin{minipage}[b]{0.49\columnwidth}
%     \centering
%     \includegraphics[width=\textwidth]{img/real_inspire.jpg}
%     \caption{\textbf{Visualization of inspire hands.}}
%     \label{realrobot}
%   \end{minipage}
%   \begin{minipage}[b]{0.49\columnwidth}
%     \centering
%     \includegraphics[width=\textwidth]{img/gello.jpg}
%     \caption{\textbf{Overview of real-world experiments setup.}}
%     \label{action}
%   \end{minipage}
%   \hfill 
%   \begin{minipage}[b]{0.49\columnwidth}
%     \centering
%     \includegraphics[width=\textwidth]{img/manus.jpg}
%     \caption{\textbf{Visualization of inspire hands.}}
%     \label{realrobot}
%   \end{minipage}
% \end{figure}

We collect 10 trajectories with each object for post-training. The specific hyper-parameters for real world experiments are shown in the TABLE \ref{tab:hyperparams-real}.

\begin{table}[!h]
\centering
\renewcommand{\arraystretch}{1.1}
\caption{Hyper-parameters for post-training experiments in real world.}
\vspace{-2mm}
\label{tab:hyperparams-real}
\small
\begin{tabular}{lc}
\toprule
\textbf{Parameter} & \textbf{Value} \\ 
\midrule
Input Image Size & $224 \times 224$ \\
Training Iterations & 30k \\
Batch Size & 32 \\
Learning Rate & $2 \times 10^{-5}$ \\
Weight Decay & 0.05 \\
Optimizer & AdamW \\
Precision & BF16 \\
Diffusion Model Objective & $x_0$-prediction \\
Selection Scale & 2 \\ 
Noise Schedule & Cosine \\
Noise decay & 0.5 \\
Window numbers & 4 \\
Sampling steps & 50 \\
Jumpy Denoising Interval & 10 \\
\bottomrule
\end{tabular}
\end{table}

\subsection{Related Work}
\label{relatedwork}
\subsubsection{\textbf{Vision-Language-Action Models}}
Developing versatile and robust robot controllers based on VLA models has emerged as a cornerstone of contemporary robotics research~\cite{zitkovich2023rt, ghosh2024octo, chen2025conrft}. By pre-training on large-scale, heterogeneous robotic datasets, VLA models~\cite{pi05, gr00t} achieve a profound synthesis of visual perception~\cite{qu2025spatialvla}, semantic reasoning~\cite{zhao2025cot}, and action generation~\cite{liang2025discrete}. These models serve as powerful backbones that significantly enhance robotic policies, particularly in handling task complexity~\cite{deng2025graspvla} and long-horizon reasoning~\cite{zhangdreamvla}. However, the inherent heterogeneity of robots often prevents pre-trained VLAs from attaining optimal performance on specific downstream tasks without post-training~\cite{gr00t}. To bridge this gap, researchers often inject embodiment-specific priors during the post-training stage to enhance structural understanding, such as 3D spatial grounding~\cite{qu2025spatialvla}, instruction-following~\cite{xvla}, historical reasoning~\cite{shi2025memoryvla} and so on. Despite these advancements, existing post-training paradigms are largely confined to single-embodiment tasks, which severely restricts their generalization and transferability across different embodiments~\cite{fan2025xr}. While recent researchers have attempted to mitigate this limitation by using latent unified action representations~\cite{univla} or prompt-based learning concepts~\cite{xvla} in pre-training stage, these approaches still rely on separate action heads tailored to specific downstream tasks. Motivated by the need for a robust architectural framework that effectively addresses the multi-embodiment post-training, we present X-DiffVLA in this paper.

\subsubsection{\textbf{Diffusion Generation Models for Multimodal Distributions}}
Diffusion-based generative models~\cite{ho2020denoising} have recently emerged as a cornerstone in various domains, achieving substantial breakthroughs in speech synthesis~\cite{huang2022fastdiff}, video generation~\cite{ho2022video}, and robotic control~\cite{chi2025diffusion}. Their primary advantage lies in their superior capacity for the comprehension and reconstruction of complex multimodal data~\cite{croitoru2023diffusion}. Within this context, utilizing the diffusion denoising process via conditional signals to target distribution has become a critical area of investigation. Diffuser~\cite{janner2022planning} models both states and actions simultaneously, allowing for reward-based classifier guidance; however, it demonstrates limited robustness. Similarly, Diffuse-CloC~\cite{huang2025diffuse} models the joint distribution of actions and achieves steerable generation by conditioning on predicted states. Leveraging the superior conditional generation capabilities of diffusion models in multimodal reconstruction, X-DiffVLA adopts the diffusion head to facilitate robust cross-embodied post-training.

\subsection{Visualization}

\begin{figure}[h]
  \centering
  \begin{subfigure}[b]{\columnwidth}
    \centering
    \includegraphics[width=\textwidth]{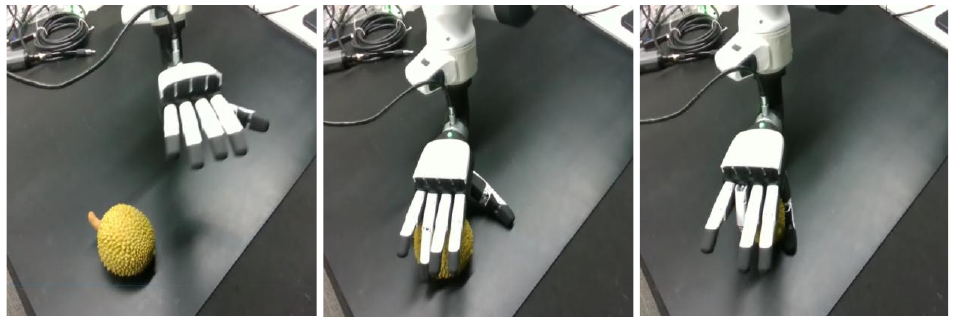}
    \caption{Suboptimal transfer from the panda gripper to dexterous inspire hands.}
    \vspace{2mm}
    \label{poor}
  \end{subfigure}
  \vspace{2mm} % 上下行之间的间距
  \begin{subfigure}[b]{\columnwidth}
    \centering
    \includegraphics[width=\textwidth]{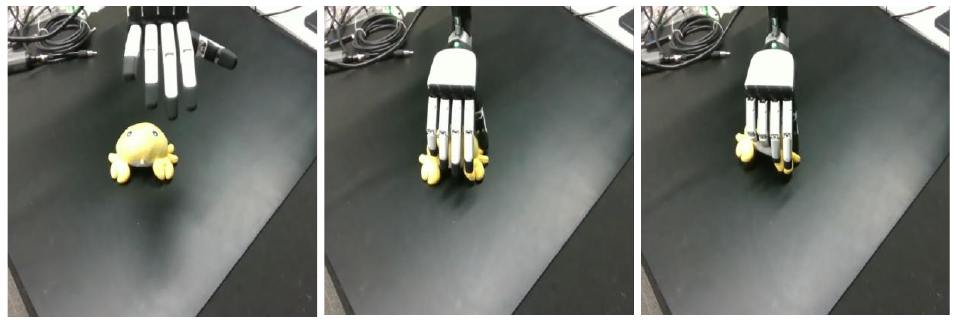}
    \caption{Successful cross-embodiment transfer with basic functional alignment.}
    \label{middle}
  \end{subfigure}
  \vspace{2mm} % 上下行之间的间距
  \begin{subfigure}[b]{\columnwidth}
    \centering
    \includegraphics[width=\textwidth]{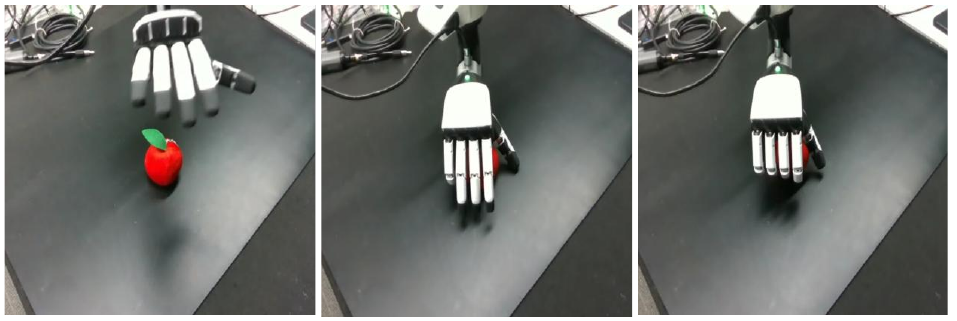}
    \caption{Another successful knowledge transfer case.}
    \label{good}
  \end{subfigure}
  \caption{\textbf{Visualization of real-world knowledge transfer across different embodiments.}}
  \label{transfer}
\end{figure}

We provide an extensive visualization of cross-embodied knowledge transfer in real-world scenarios. Fig. \ref{poor} to \ref{good} collectively demonstrate how the dexterous grasping policy leverages data from parallel grippers. Specifically, the inspire hand first engages the object with distal fingers, which is encouraged by gripper data, before activating additional digits to achieve a stable, robust grasp. However, a performance gap in transfer quality remains evident. Consequently, our future work will focus on modulating the temporal dynamics of MPTD within the X-DiffVLA framework, enabling the dynamic allocation of cross-embodied knowledge transfer intensity based on the real-time progression of the task.

\end{document}